%% file: main.tex
\documentclass[11pt, intlimits]{article}
\usepackage[letterpaper, margin=1.0in, footskip=30pt]{geometry}

\usepackage{amsmath}
\usepackage[T1]{fontenc}
\usepackage{mathrsfs}
\usepackage{graphicx}
\usepackage{pgfplots}
\usepackage{float}
\usepackage{tabu}
\usepackage{booktabs}
\usepackage{tikz}
\usepackage{tikz-cd}
\usepackage{url}
\graphicspath{{graphics/}}

\usepackage[backend=biber, sorting=none, style=numeric-comp]{biblatex}
\usepackage{caption}
\begin{document}

\thispagestyle{empty}

\begin{center}
{\LARGE Some Best Practices in Operator Learning}
\\
25 November 2024
\end{center}

\noindent
\begin{minipage}[t]{0.5\textwidth}
\begin{center}
Dustin Enyeart \\
Department of Mathematics \\
Purdue University \\
\url{denyear@purdue.edu} \\
\end{center}
\end{minipage}%
\hfill
\begin{minipage}[t]{0.5\textwidth}
\begin{center}
Guang Lin\footnotemark \\
Department of Mathematics \\ 
School of Mechanical Engineering \\
Purdue University \\
\url{guanglin@purdue.edu} 
\end{center}
\end{minipage}
\footnotetext{Corresponding author}

\begin{abstract}
\noindent \input{abstract} 
\newline 
\newline 
\textbf{Keywords:} \input{keywords}
\end{abstract}

\section{Introduction}
\input{introduction}

\section{Architectures}
\input{architectures}

\section{Differential Equations}
\input{diffeq}

\section{Activation Function}
\input{act-fns.tex}

\section{Dropout}
\input{dropout}

\section{Stochastic Weight Averaging}
\input{swa}

\section{Learning Rate Finder}
\input{lr-finder}

\section{Conclusion}
\input{conclusion}

\section*{Code Availability}
\input{code.tex}

\section*{Declaration of Competing Interest}
\input{competing-interest.tex}

\section*{Acknowledgment}
GL would like to thank the support of National Science Foundation (DMS-2053746, DMS-2134209, ECCS-2328241, CBET-2347401 and OAC-2311848), and U.S.~Department of Energy (DOE) Office of Science Advanced Scientific Computing Research program DE-SC0023161, and DOE–Fusion Energy Science, under grant number: DE-SC0024583.

\printbibliography

\end{document}

%% file: abstract.tex
Hyperparameters searches are computationally expensive. 
This paper studies some general choices of hyperparameters and training methods specifically for operator learning. 
It considers the architectures DeepONets, Fourier neural operators and Koopman autoencoders for several differential equations to find robust trends. 
Some options considered are activation functions, dropout and stochastic weight averaging. 

%% file: keywords.tex
Operator learning, Best practices, Hyperparameters, Training methods, DeepONet, Fourier neural operator, Koopman autoencoder, Activation function, Dropout, Stochastic weight averaging, Learning rate finder

%% file: introduction.tex
A \emph{neural operator}\index{neural operator} is a neural network that is intended to approximate an operator between function spaces \cite{kovachki2024operator, boulle2023mathematical, winovich2021neural}. 
An example of an output function for a neural operator is a solution to a differential equation. 
Examples of input functions for a neural operator are the initial conditions or the boundary conditions for the differential equation.
The study of neural operators is called \emph{operator learning}\index{operator learning}.

Various choices of hyperparameters and training strategies are made when training a neural network.
Limiting such choices to known robust choices can significantly improve the speed of hyperparameters searches. 
This paper studies the effect of some of these choices in operator learning, that is, it tries to find some best practices in operator learning. 
This is not a comparison between different architectures. 
In order, this paper studies the choice of activation function, the use of dropout, the use of stochastic weight averaging and the use of a learning rate finder.

%% file: architectures.tex
This section introduces the architectures used in this paper. 
These are DeepONets, Fourier neural operators and Koopman autoencoders.

\subsection{DeepONets}

\emph{Deep neural operators}\index{deep neural operator}, which are abbreviated as \emph{DeepONets}\index{DeepONet}, are a neural operator architecture \cite{DeepONet, lanthaler2022error, goswami2022physics, he2023novel, cho2024learning, lu2022comprehensive}. 
They consist of two parts. 
The first part encodes information about the differential equation into a latent space.
It is called the \emph{branch network}\index{branch network}.
The second part encodes a position into a latent space.
It is called the \emph{trunk network}\index{trunk network}.
An example input for the branch network is the initial condition for a dynamic differential equation, and an example for the trunk network is the space-time point to evaluate the solution to the differential equation.
In practice, there is usually only one forward pass for the branch network, but many forward passes for the trunk network to encode all the desired positions.

The dimensions of the latent spaces of the branch network and the trunk network are the same. 
And, the output of such a model is the dot product between both latent spaces. 
Intuitively, the branch network encodes the solution to the differential equation as a set of basis functions, and the trunk network encodes the position as the coefficients of these basis functions.

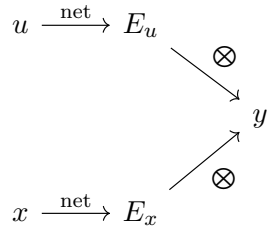
\begin{figure}[H]
\[
\begin{tikzcd}
    u \arrow{r}{\mathrm{net}} & E_{u} \arrow{dr}{\bigotimes} & 
    \\
    & & y
    \\
    x \arrow{r}{\mathrm{net}} & E_{x} \arrow[swap]{ur}{\bigotimes} & 
    \\
\end{tikzcd}
\]
\caption{The DeepONet architecture:
    The input $u$ is the input function, and the input $x$ is the point where the output function is evaluated. Their encodings are denoted by $E_u$ and $E_x$, respectively. 
    The output is denoted by $y$.}
\end{figure}

\subsection{Fourier Operator Networks}

\emph{Fourier neural operators}\index{Fourier neural operator}, which are abbreviated as \emph{FNO}\index{FNO}, are a neural operator architecture \cite{li2020fourier, qin2024toward, li2023fourier, brandstetter2022lie, kovachki2021universal, tran2021factorized, tancik2020fourier, lu2022comprehensive}. 
They are based on spectral convolution layers. 

In a \emph{spectral convolution layer}\index{spectral convolution layer}, first the Fourier transform is applied. 
Then, the higher modes are dropped. 
Then, a linear function is applied. 
Then, the inverse Fourier transform is applied \cite{stein2011fourier}.
Similarly to residuals for a normal convolution layer, a linear function is applied to the original input, and the output is added elementwise to the output of the inverse Fourier transform. 
Finally, an activation function is applied. 

While a normal convolution layer is local, a spectral convolution layer is global.
Furthermore, for a spectral convolution layer to be computationally reasonable, the fast Fourier transform needs to be used \cite{rao2011fast}. 
Thus, the data needs to be structured as a grid. 

First, in a Fourier neural operator, a feed-forward neural network is applied to increase the channel dimension. 
Then, a sequence of spectral convolution layers are applied. 
Then, a feed-forward neural network is applied to decrease the channel dimension to the output size.

The original channels include the input function, such as the initial conditions.
For time-independent differential equation, the Fourier neural network can be applied for each time step, or the desired time can be included in the channels.
In this paper, the time is included in the channels, that is, for each desired time, the Fourier neural network is applied to the initial conditions together with the desired time.  
When using grids of various resolutions, the spatial step size is also included. 
And, for inhomogeneous equations, the spatial position is also included.

\begin{figure}[H]
\[
\begin{tikzcd}
    x \arrow{r}{\mathrm{net}} & \arrow{r}{\mathrm{spectral}} & \cdots \arrow{r}{\mathrm{spectral}} & \arrow{r}{\mathrm{net}} & y
    \\
\end{tikzcd}
\]
\caption{
    A Fourier neural operator: The input is denoted by $x$. First, a feed-forward neural network is used to increase the channel dimension. Then, a sequence of spectral convolution layers are applied. Then, a feed-forward neural network is used to decrease the channel dimension. The output is denoted by $y$.
}
\end{figure}

\begin{figure}[H]
\[
\begin{tikzcd}
    x \arrow{r}{\mathrm{FFT}} \arrow[swap]{drr}{\mathrm{linear}}
    & \arrow{r}{\mathrm{filter}}
    & \arrow{r}{\mathrm{linear}}
    & \arrow{r}{\mathrm{IFFT}}
    & \arrow{r}{\bigoplus}
    & \arrow{r}{\mathrm{act}} & y
    \\
    & & \arrow[swap]{urrr}{\bigoplus} & & & 
    \\
\end{tikzcd}
\]
\caption{
    A spectral convolution layer: 
    The input is denoted by $x$. 
    On the bottom, a linear layer is applied to each channel. 
    On the top, first the fast Fourier transform is applied. 
    Then, the higher modes are dropped. 
    Then, a linear layer is applied to each channel. 
    Then, the inverse fast Fourier transform is applied.
    Then, the top and bottom are added elementwise, and an activation function is applied. 
    The output is denoted by $y$.
}
\end{figure}
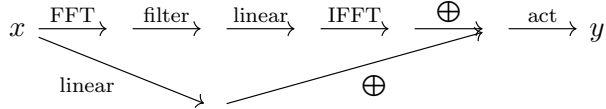

\subsection{Koopman Autoencoders}

\emph{Koopman autoencoders}\index{Koopman autoencoder} are a neural operator architecture that are used for time-dependent differential equations \cite{lusch2018deep, koop1, mamakoukas2020learning, huang2020data, klus2020data}. 
They are popular\index{applications of Koopman autoencoders} for dynamic mode decomposition\index{dynamic mode decomposition} \cite{dymodedecomp1, dymodedecomp2, kutz2016multiresolution, takeishi2017learning, bagheri2013koopman} and control\index{control} \cite{budivsic2020koopman, li2019learning, kaiser2020data, han2020deep, bruder2021koopman, bruder2019modeling, arbabi2018data, abraham2017model}. 

The Koopman formulation of classical mechanics\index{Koopman formulation of mechanics} is an alternative framework of classical mechanics \cite{koopman1931hamiltonian, brunton2021modern, bruce2019koopman}. 
It was inspired by the Hamiltonian formulation of quantum mechanics \cite{morrison1990understanding, ohanian1989principles, griffiths2018introduction}. 
In this theory, the physical state\index{physical space} at a given time is represented as a state in an infinite-dimensional latent space\index{latent space}, and an infinite-dimensional operator governs the time evolution in this latent space.
Intuitively, the Koopman formulation can be thought of as removing nonlinearity by infinitely increasing the dimension. 

The Koopman formulation of classical mechanics can be discretized\index{discretized Koopman formulation} to provide a numerical scheme by approximating the time-evolution operator by a finite-dimensional matrix $K$.
In this scheme, a physical state $s_0$ can be evolved into a later physical state $s_n$ by encoding it into the latent space, applying the matrix $K$ repetitively and then decoding it back to the physical space, that is, the equation 
\[
    s_n = R \circ K^{n} \circ E(s_0)
\]
approximately holds, where $E$ is a discretized encoder and $R$ is a discretized decoder.
In this numerical scheme, the dimension of this latent space is called the \emph{encoding dimension}\index{encoding dimension}.

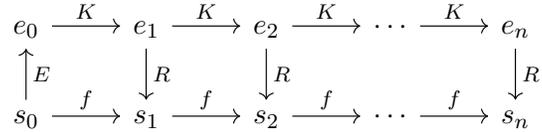
\begin{figure}[H]
\[
\begin{tikzcd}
    e_0 \arrow{r}{K}
    & e_1 \arrow{d}{R} \arrow{r}{K}
    & e_2 \arrow{d}{R} \arrow{r}{K}
    & \cdots \arrow{r}{K}
    & e_n \arrow{d}{R} \\
    s_0 \arrow[swap]{u}{E} \arrow{r}{f}
    & s_1 \arrow{r}{f}
    & s_2 \arrow{r}{f}
    & \cdots \arrow{r}{f}
    & s_n \\
\end{tikzcd}
\]
\caption{Discretization of the Koopman formulation into a numerical scheme: The physical states at successive time points are denoted by $s_0$, $s_1$, $\dots$, $s_{n-1}$ and $s_n$. The encoded states at successive time points are denoted by $e_0$, $e_1$, $\dots$, $e_{n-1}$ and $e_n$. The function $f$ is the true time evolution of the physical state by the time step. The discretized Koopman operator, encoder and decoder are denoted by $K$, $E$ and $R$, respectively.}
\end{figure}    

\emph{Koopman autoencoders}\index{Koopman autoencoder}\index{Koopman architecture}\index{Koopman neural network} are based on the discretized Koopman formulation where the operator\index{Koopman operator}, encoder\index{Koopman encoder} and decoder\index{Koopman decoder} are neural networks.
The encoder and decoder normally consists of fully connected layers or convolution layers. 
And, the operator is a fully connected layer without a bias, and an activation function\index{activation function} is not used between successive applications of it.  
Because the operator is applied repeatedly, it can be useful to use gradient clipping. 

The loss for DeepONets and Fourier neural operators is the mean squared error between the output of the model and the true solution.
For Koopman autoencoders, it is better to use additional loss terms during training \cite{enyeart2024loss}. 
The training loss used for Koopman autoencoders in this paper is 
\[
\begin{split}
    & \frac{1}{n} \cdot \sum_i \vert \vert R \circ K^i \circ E(v_0) - v_i \vert \vert^2 \\
    & + \frac{1}{n} \cdot \sum_{i} \vert \vert R \circ E(v_i) - v_i \vert \vert^2 \\
    & + \vert \vert K \circ K^{\mathrm{T}} - I \vert \vert^2, 
\end{split}
\]
where $E$ is the encoder, $R$ is the encoder, $K$ is the Koopman operator, $n$ is the number of time steps and each $v_i$ is a vector in the physical space. 
Intuitively, the first term is the accuracy, the second term regulates the relation between the encoder and the decoder, and the last term forces the operator to be unitary. 
Furthermore, a mask is on the Koopman operator so that it is tridiagonal. 
The intuition of these choices is from the Koopman formulation of classical mechanics. 

%% file: diffeq.tex
This section introduces the differential equations that are used for the numerical experiments in this paper.
The ordinary differential equations are the equation for the pendulum, the Lorenz system and a fluid attractor equation. 
The partial differential equations are Burger's equation and the Korteweg-de-Vries equation.

\subsection{Pendulum}

The equation for the \emph{pendulum}\index{pendulum} is the differential equation
\[
    \frac{\mathrm{d}^2 \theta}{\mathrm{d}^2 t}
    = - \sin (\theta)
    . 
\]
This is a time-dependent second-order nonlinear ordinary differential equation whose dimension is $1$. 
This equation models the motion of a pendulum in a constant gravitational field, where the variable $\theta$ is the angle of the pendulum from vertical.

In the numerical experiments in this paper, the models attempt to learn the solution to this differential equation as a function of the initial condition.
To get data, initial positions are generated randomly.
Then, this differential equation is numerically solved for these initial conditions using the Runge-Kutta Method \cite{leveque2007finite}.

\subsection{Lorenz System}

The \emph{Lorenz system}\index{Lorenz system} is the differential equation
\[
\begin{cases}
    \frac{\mathrm{dx}}{\mathrm{d}t} = y - x  \\
    \frac{\mathrm{dy}}{\mathrm{d}t} = x - x \cdot z - y  \\
    \frac{\mathrm{dz}}{\mathrm{d}t} = x \cdot y - z. \\ 
\end{cases}
\]
This is a time-dependent first-order nonlinear ordinary differential equation whose dimension is $3$. 
Historically, this equation was used in weather modeling.
Now, it provides a benchmark example of a chaotic system\index{chaotic system}.

In the numerical experiments in this paper, the models attempt to learn the solution to this differential equation as a function of the initial condition.
To get data, initial positions are generated randomly.
Then, this differential equation is numerically solved for for these initial conditions using the Runge-Kutta Method \cite{leveque2007finite}.

\subsection{Fluid Attractor Equation}

The differential equation 
\[
\begin{cases}
    \frac{\mathrm{dx}}{\mathrm{d}t} = x - y + x \cdot z  \\
    \frac{\mathrm{dy}}{\mathrm{d}t} = x + y + y \cdot z \\
    \frac{\mathrm{dz}}{\mathrm{d}t} = x^2 + y^2 + z \\ 
\end{cases}
\]
is used to model fluid flow around a cylinder\index{fluid attractor equation} \cite{noack2003hierarchy}. 
It is a time-dependent first-order nonlinear ordinary differential equation whose dimension is $3$.

In the numerical experiments in this paper, the models attempt to learn the solution to this differential equation as a function of the initial condition.
To get data, initial positions are generated randomly.
Then, this differential equation is numerically solved for for these initial conditions using the Runge-Kutta Method \cite{leveque2007finite}.

\subsection{Burger's Equation}

\emph{Burger's equation}\index{Burger's equation} is the differential equation
\[
    \frac{\partial u}{\partial t}
    = \frac{\partial^2 u}{\partial^2 x} - u\frac{\partial u}{\partial x} 
    .
\]
This is a time-dependent first-order nonlinear partial differential equation whose domain dimension and range dimension are both $1$. 
It is used to model some fluids. 

In the numerical experiments in this paper, the models attempt to learn the solution to this differential equation as a function of the initial condition.
Furthermore, Dirichlet boundary conditions are used for this equation such that the value of the unknown function is $0$ on the boundary.
To get data, initial conditions are made by generating random symbolic expressions that satisfy the boundary conditions, and these expressions are then numerically sampled. 
Then, the partial differential equation is numerically solved for these initial conditions \cite{leveque1992numerical}.

\subsection{Korteweg-de-Vries Equation}

The \emph{Korteweg-de-Vries equation}\index{Korteweg-de-Vries equation}, which is abbreviated as the \emph{KdV equation}\index{KdV equation}, is the differential equation
\[
    \frac{\partial u}{\partial t}
    = 6u\frac{\partial u}{\partial x}
    - \frac{\partial^3u}{\partial ^3x}
    .
\]
This is a time-dependent first-order nonlinear partial differential equation whose domain dimension and range dimension are both $1$. 
It is used to model some fluids. 

In the numerical experiments in this paper, the models attempt to learn the solution to this differential equation as a function of the initial condition.
Furthermore, periodic boundary conditions are used for this equation, that is, the value of the unknown function is the same on each endpoint of the spatial domain. 
To get data, initial conditions are made by generating random symbolic expressions that satisfy the boundary conditions, and these expressions are then numerically sampled. 
Then, the partial differential equation is numerically solved for these initial conditions \cite{zabusky1965interaction}.

%% file: act-fns.tex
This section studies the use of different activation functions in operator learning. 
The activation functions that are compared are the hyperbolic tangent function, the rectified linear unit, the gaussian error linear unit and the exponential linear unit.  
The \emph{rectified linear unit}\index{rectified linear unit} is the function
\[
    \max(0, x),
\]
and it is denoted as $\mathrm{relu}(x)$.
The \emph{gaussian error linear unit}\index{gaussian error linear unit} is the function 
\[
    x \cdot \Theta(x), 
\]
where $\Theta(x)$ is the cumulative distribution function of the standard normal distribution, that is, the function 
\[
    \frac{1}{2} \cdot 
    \left(
        1 + \frac{2}{\sqrt{\pi}} \cdot \int_{0}^{x} \exp(-t^2) \mathrm{d}t
    \right). 
\]
The gaussian error linear unit is denoted as $\mathrm{gelu}(x)$. 
The \emph{exponential linear unit}\index{exponential linear unit} is the function
\[
    \begin{cases}
        x & \text{if } x > 0, \\
        \exp(x) - 1 & \text{if } x \leq 0,
    \end{cases}
\]
and it is denoted as $\mathrm{elu}(x)$.

\begin{figure}[H]
\begin{center}
    \includegraphics[scale=.35]{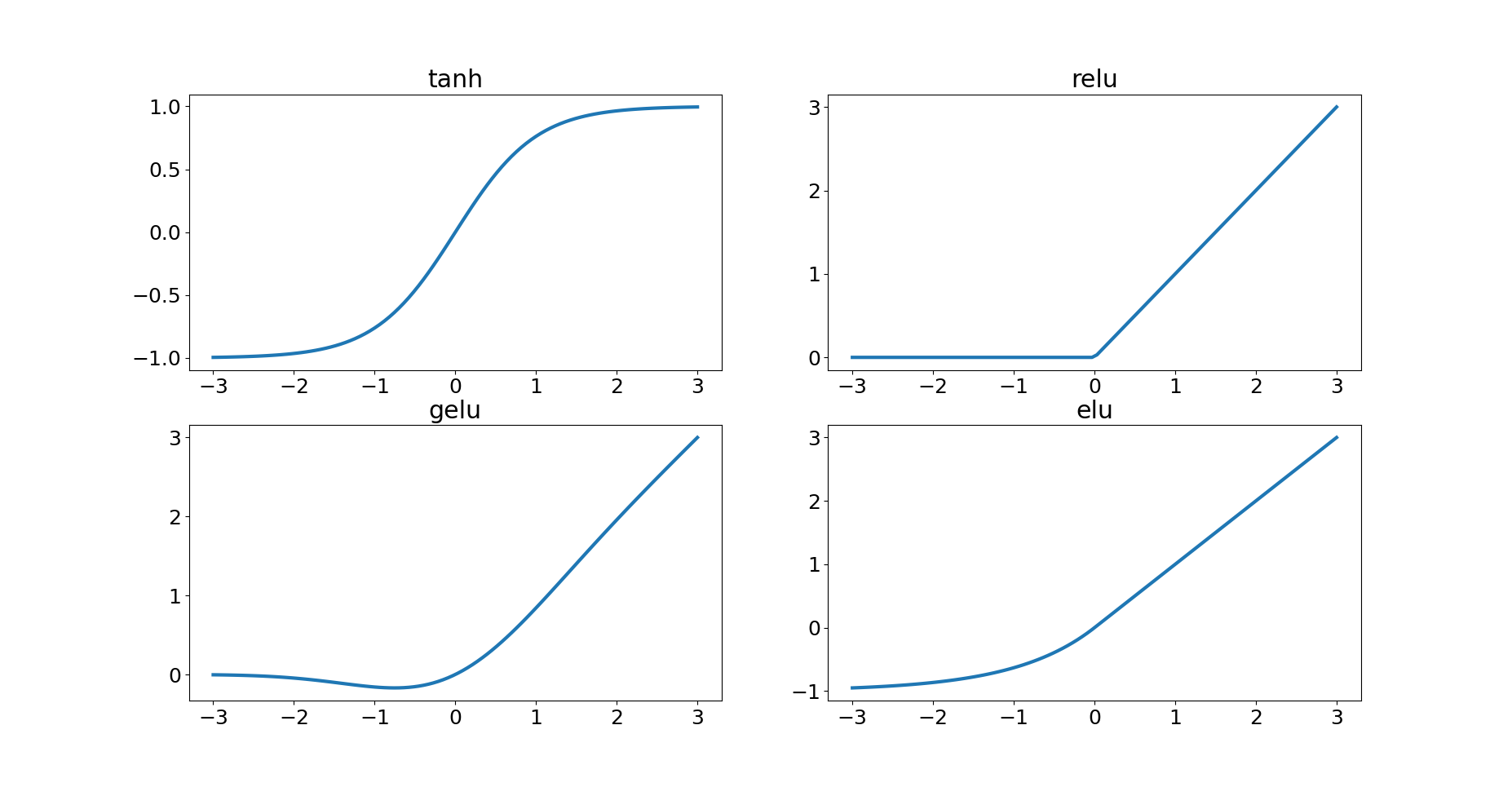}
    \caption{Activation functions:
        The upper left is the hyperbolic tangent. 
        The upper right is the rectified linear unit.
        The lower left is the gaussian error linear unit.
        The lower right is the exponential linear unit.
        }
\end{center}
\end{figure}

Intuitively, in operator learning, the activation functions should be smooth because they are normally approximating smooth functions. 
Furthermore, the activation functions $\tanh$ and $\mathrm{gelu}$ are the most common in practice.

Numerical experiments were done for three architectures and each for two differential equations. 
DeepONets were used for the Lorenz system and Burger's equation. 
Fourier neural operators were used for Burger's equation and the KdV equation. And, Koopman autoencoders were used for the equation for the pendulum and the fluid attractor equation.
The results are in Table \ref{actfn-deeponet}, Table \ref{actfn-fno} and Table \ref{actfn-koopman}, respectively.

\begin{table}[H]
    \caption{Results for activation functions for DeepONets: The left table is for the Lorenz system, and the right table is for Burger's equation.}
    \label{actfn-deeponet}
    \begin{minipage}{.5\textwidth}
        \vspace{1em}
        \begin{center}
        Lorenz \\
        \begin{tabular}{lr}
            error & activation \\
            \toprule
            1.388e-2 & relu \\
            5.429e-3 & gelu \\
            1.446e-2 & elu \\
            6.608e-3 & tanh \\
        \end{tabular}
        \end{center}
    \end{minipage}
    \hspace{1cm}
    \begin{minipage}{.5\textwidth}
        \vspace{1em}
        \begin{center}
        Burger's equation \\
        \begin{tabular}{lr}
            error & activation \\
            \toprule
            3.424e-3 & relu \\
            2.576e-3 & gelu \\
            4.183e-3 & elu \\
            3.363e-3 & tanh \\
        \end{tabular}
        \end{center}
    \end{minipage}
\end{table}

\begin{table}[H]
    \caption{Results for activation functions for Fourier neural operators: The left table is for Burger's equation, and the right table is for the KdV equation.}
    \label{actfn-fno}
    \begin{minipage}{.5\textwidth}
        \vspace{1em}
        \begin{center}
        Burger's \\
        \begin{tabular}{lr}
            error & activation \\
            \toprule
            6.428e-4 & relu \\
            5.013e-4 & gelu \\
            5.355e-4 & elu \\
            1.777e-3 & tanh \\
        \end{tabular}
        \end{center}
    \end{minipage}
    \hspace{1cm}
    \begin{minipage}{.5\textwidth}
        \vspace{1em}
        \begin{center}
        KdV \\
        \begin{tabular}{lr}
            error & activation \\
            \toprule
            4.116e-3 & relu \\
            3.731e-3 & gelu \\
            3.909e-3 & elu \\
            6.220e-3 & tanh \\
        \end{tabular}
        \end{center}
    \end{minipage}
\end{table}

\begin{table}[H]
    \caption{Results for activation functions for Koopman autoencoders: The left table is for the equation for the pendulum, and the right table is for the fluid attractor equation.}
    \label{actfn-koopman}
    \begin{minipage}{.5\textwidth}
        \vspace{1em}
        \begin{center}
        pendulum \\
        \begin{tabular}{lr}
            error & activation \\
            \toprule
            7.719e-4 & relu \\
            1.752e-4 & gelu \\
            1.360e-4 & elu \\
            8.705e-4 & tanh \\
        \end{tabular}
        \end{center}
    \end{minipage}
    \hspace{1cm}
    \begin{minipage}{.5\textwidth}
        \vspace{1em}
        \begin{center}
        fluid attractor \\
        \begin{tabular}{lr}
            error & activation \\
            \toprule
            6.900e-4 & relu \\
            1.112e-5 & gelu \\
            3.322e-5 & elu \\
            4.602e-5 & tanh \\
        \end{tabular}
        \end{center}
    \end{minipage}
\end{table}

\begin{figure}[H]
\begin{center}
    \includegraphics[scale=.5]{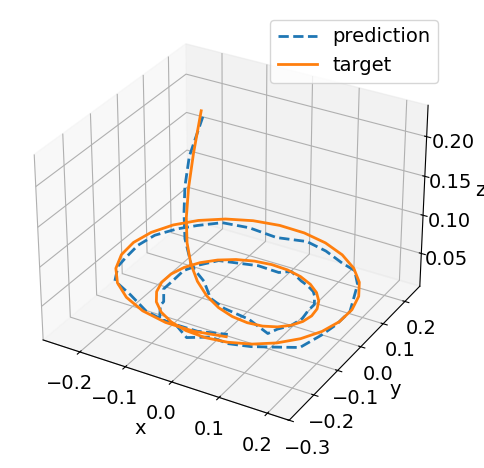}
    \includegraphics[scale=.5]{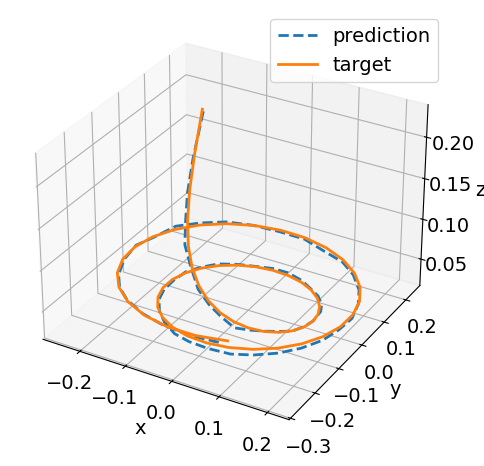}
    \caption{Progress for different activation functions: Both plots show the trajectories for target and the prediction of a Koopman autoencoder for the fluid attractor equation at 1000 epochs. On the left, the activation function used is $\mathrm{relu}$. On the right, the activation function used is $\mathrm{gelu}$. The prediction on the right is visibly more accurate.}
\end{center}
\end{figure}

In every experiment, the activation function $\mathrm{gelu}$ preformed the best.
Thus, is is recommend to use the activation function $\mathrm{gelu}$ in operator learning.
In three out of the six experiments, the activation function $\mathrm{relu}$ preformed the worst, which is intuitively expected because it is not differentiable. 
In three out of the six experiments, the activation function $\tanh$ preformed the worst, which is surprising because it is often used in practice.

%% file: dropout.tex
This section studies the use of dropout in operator learning.
In a layer in a neural network, \emph{dropout}\index{dropout} is a training technique where the entries of the matrix and displacement vector of a linear layer are randomly set to $0$ \cite{hinton2012improving, srivastava2014dropout}.
The \emph{dropout ;}\index{dropout rate} is the prescribed fraction of entries to set to $0$.
Dropout often prevents neural networks from overfitting during training. 

Numerical experiments were done for three architectures and each for two differential equations. 
DeepONets were used for the Lorenz system and Burger's equation. 
Fourier neural operators were used for Burger's equation and the KdV equation.
And, Koopman autoencoders were used for the equation for the pendulum and the fluid attractor equation.
The results are in Table \ref{dropout-deeponet}, Table \ref{dropout-fno} and Table \ref{dropout-koopman}, respectively.
For the Koopman autoencoders, dropout was not experimented with for the operators, but only for the encoders and decoders.

\begin{table}[H]
    \caption{Results for dropout for DeepONets: The left table is for the Lorenz system, and the right table is for Burger's equation.}
    \label{dropout-deeponet}
    \begin{minipage}{.5\textwidth}
        \vspace{1em}
        \begin{center}
        Lorenz \\
        \begin{tabular}{ll}
        error & dropout \\
        \toprule
        6.999e-3 & 0.0 \\
        3.363e-2 & 0.05 \\
        5.396e-2 & 0.1 \\
        7.036e-2 & 0.15 \\
        \end{tabular}
        \end{center}
    \end{minipage}
    \hspace{1cm}
    \begin{minipage}{.5\textwidth}
        \vspace{1em}
        \begin{center}
        Burger's \\
        \begin{tabular}{lr}
        error & dropout \\
        \toprule
        3.211e-3 & 0.0 \\
        7.125e-3 & 0.05 \\
        8.874e-3 & 0.1 \\
        1.036e-2 & 0.15 \\
        \end{tabular}
        \end{center}
    \end{minipage}
\end{table}

\begin{table}[H]
    \caption{Results for dropout for Fourier neural operators: The left table is for the Burger's equation, and the right table is for the KdV equation.}
    \label{dropout-fno}
    \begin{minipage}{.5\textwidth}
        \vspace{1em}
        \begin{center}
        Burger's \\
        \begin{tabular}{ll}
        error & dropout \\
        \toprule
        7.585e-4 & 0.0 \\
        1.138e-3 & 0.05 \\
        1.126e-3 & 0.1 \\
        1.958e-3 & 0.15 \\
        \end{tabular}
        \end{center}
    \end{minipage}
    \hspace{1cm}
    \begin{minipage}{.5\textwidth}
        \vspace{1em}
        \begin{center}
        KdV \\
        \begin{tabular}{lr}
        error & dropout \\
        \toprule
        3.090e-3 & 0.0 \\
        4.709e-3 & 0.05 \\
        3.584e-3 & 0.1 \\
        4.286e-3 & 0.15 \\
        \end{tabular}
        \end{center}
    \end{minipage}
\end{table}

\begin{table}[H]
    \caption{Results for dropout for Koopman autoencoders: The left table is for the equation for the pendulum, and the right table is for the fluid attractor equation.}
    \label{dropout-koopman}
    \begin{minipage}{.5\textwidth}
        \vspace{1em}
        \begin{center}
        pendulum \\
        \begin{tabular}{lr}
            error & dropout \\
            \toprule
            1.325e-3 & 0.0 \\
            1.385e-2 & 0.05 \\
            2.204e-2 & 0.1 \\
            3.369e-2 & 0.15 \\
        \end{tabular}
        \end{center}
    \end{minipage}
    \hspace{1cm}
    \begin{minipage}{.5\textwidth}
        \vspace{1em}
        \begin{center}
        fluid attractor \\
        \begin{tabular}{ll}
            error & dropout \\
            \toprule
            4.696e-5 & 0.0 \\
            2.238e-4 & 0.05 \\
            3.336e-4 & 0.1 \\
            4.982e-4 & 0.15 \\
        \end{tabular}
        \end{center}
    \end{minipage}
\end{table}

\begin{figure}
\begin{center}
    \includegraphics[scale=.7]{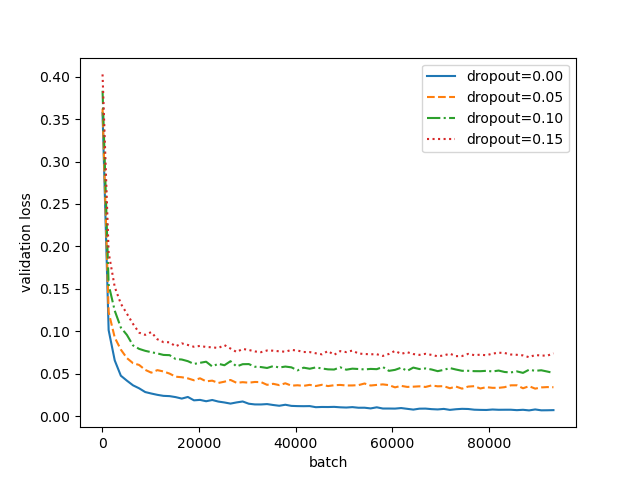}
    \caption{Validation loss using dropout: A DeepONet is used for the Lorenz system with various dropout rates. 
        }
    \label{dropout-val-loss}
\end{center}
\end{figure}

In every experiment, dropout decreased the accuracy of the model. 
Furthermore, Figure \ref{dropout-val-loss} displays the validation loss for a DeepONet for the Lorenz system with various dropout rates, and it demonstrates that increasing the dropout rate decreases the accuracy.
Thus, it is recommended to not use dropout in operator learning. 

%% file: swa.tex
This section studies the use of stochastic weight averaging in operator learning.
\emph{Stochastic weight averaging}\index{stochastic weight averaging} is an addition to the optimizer that averages the parameters of a neural networks at different points during training that often improves convergence and prevents overfitting \cite{izmailov2018averaging, athiwaratkun2018there, nikishin2018improving, pytorch_swa}. 
For stochastic weight averaging, the learning rate strategy is applied normally for the majority of the epochs. 
Then, during the remaining part, a constant or periodic learning rate is used.  
The weights during this remaining part are averaged to get the final weights.
Normally, gradient descent will tend find minima around the edge of a flat region, and stochastic gradient descent will tend to find minima in the center of such a flat region, which, intuitively, will generalize more to unseen data.

\begin{figure}[H]
\begin{center}
    \includegraphics[scale=.15]{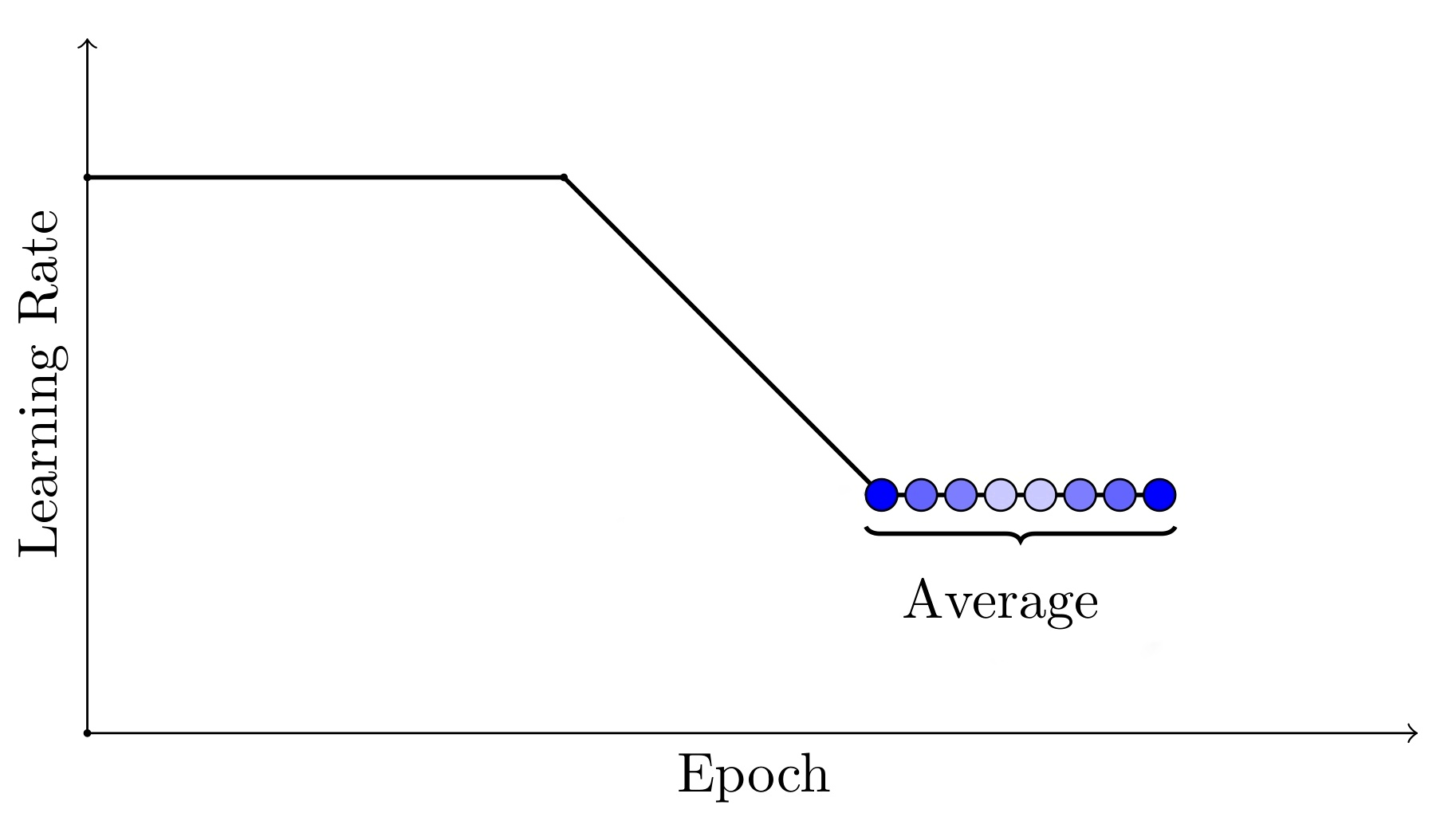}
    \caption{Stochastic weight averaging:
        The optimizer dictates the learning rate for the majority of the epochs. Towards the end of training, the learning rate is fixed. The final weights of the model are the average of the weights during this latter stage. 
        }
\end{center}
\end{figure}

Numerical experiments were done for three architectures and each for two differential equations. 
DeepONets were used for the Lorenz system and Burger's equation. 
Fourier neural operators were used for Burger's equation and the KdV equation.
And, Koopman autoencoders were used for the equation for the pendulum and the fluid attractor equation.
The results are in Table \ref{swa-deeponet}, Table \ref{swa-fno} and Table \ref{swa-koopman}, respectively.
For each model, the base learning rate was $0.001$.

\begin{table}[H]
    \caption{Results stochastic weight averaging for DeepONets: The left table is for the Lorenz system, and the right table is for the KdV equation. The base learning rate of both models was $0.001$.}
    \label{swa-deeponet}
    \begin{minipage}{.5\textwidth}
        \vspace{1em}
        \begin{center}
        Lorenz \\
            \begin{tabular}{ll}
            error & SWA LR \\
            \toprule
            1.084e-2 & none \\
            1.045e-2 & 1e-4 \\
            8.323e-3 & 1e-3 \\
            1.673e-2 & 1e-2 \\
            1.699 & 1e-1 \\
            \end{tabular}
        \end{center}
    \end{minipage}
    \hspace{1cm}
    \begin{minipage}{.5\textwidth}
        \vspace{1em}
        \begin{center}
        KdV \\
        \begin{tabular}{ll}
            error & SWA LR \\
            \toprule
            1.261e-2 & none \\
            4.700e-2 & 1e-4 \\
            1.072e-2 & 1e-3 \\
            1.910e-1 & 1e-2 \\
            9.269e-1 & 1e-1 \\
            \end{tabular}
        \end{center}
    \end{minipage}
\end{table}

\begin{table}[H]
    \caption{Results stochastic weight averaging for Fourier neural operators: The left table is for the equation for burger's equation, and the right table is for the KdV equation. The base learning rate of both models was $0.001$.}
    \label{swa-fno}
    \begin{minipage}{.5\textwidth}
        \vspace{1em}
        \begin{center}
        Burger's \\
        \begin{tabular}{ll}
            error & SWA LR \\
            \toprule
            7.973e-4 & none \\
            4.011e-4 & 1e-4 \\
            4.856e-4 & 1e-3 \\
            2.540e-1 & 1e-2 \\
            2.684e-1 & 1e-1 \\
        \end{tabular}
        \end{center}
    \end{minipage}
    \hspace{1cm}
    \begin{minipage}{.5\textwidth}
        \vspace{1em}
        \begin{center}
        KdV \\
        \begin{tabular}{ll}
            error & SWA LR \\
            \toprule
            7.34086e-3 & none \\
            4.36237e-3 & 1e-4 \\
            6.49731e-3 & 1e-3 \\
            2.07142e-1 & 1e-2 \\
            2.40617e-1 & 1e-1 \\
        \end{tabular}
        \end{center}
    \end{minipage}
\end{table}

\begin{table}[H]
    \caption{Results stochastic weight averaging for Koopman autoencoders: The left table is for the equation for the pendulum, and the right table is for the fluid attractor equation. The base learning rate of both models was $0.001$.}
    \label{swa-koopman}
    \begin{minipage}{.5\textwidth}
        \vspace{1em}
        \begin{center}
        pendulum \\
        \begin{tabular}{lr}
            error & SWA LR \\
            \toprule
            4.370e-4 & none \\
            2.122e-4 & 1e-4 \\
            4.919e-4 & 1e-3 \\
            2.536e-3 & 1e-2 \\
            3.665e-1 & 1e-1 \\
        \end{tabular}
        \end{center}
    \end{minipage}
    \hspace{1cm}
    \begin{minipage}{.5\textwidth}
        \vspace{1em}
        \begin{center}
        fluid attractor \\
        \begin{tabular}{lr}
            error & SWA LR \\
            \toprule
            4.595e-5 & none \\
            9.073e-6 & 1e-4 \\
            1.162e-5 & 1e-3 \\
            7.259e-5 & 1e-2 \\
            2.757e-2 & 1e-1 \\
        \end{tabular}
        \end{center}
    \end{minipage}
\end{table}

In each experiment, stochastic weight averaging improved the accuracy of the model when the learning rate was the same or one tenth of the original learning rate.
And, it is recommend to use stochastic weight averaging with such a learning rate. 
When the learning rate was significantly bigger than the original learning rate, the model preformed worse. 
This is common for stochastic weight averaging. 
Intuitively, if the learning rate is too big, the model weights will jump from local minimum to local minimum instead of looping around the flat region of a single local minimum, and the average of the weights will then be away from any local minimum instead of the center of a flat region. 

%% file: lr-finder.tex
A \emph{learning rate finder}\index{learning rate finder} tries to find the optimal learning automatically instead of specifying it manually \cite{smith2017cyclical, liu2019variance, gotmare2018closer}. 
This is done by sweeping through a range of learning rates and then choosing the one corresponding to the greatest descent in the loss. 
This can be beneficial because it is one less hyperparameter to optimize.

\begin{figure}[H]
\begin{center}
    \includegraphics[scale=.7]{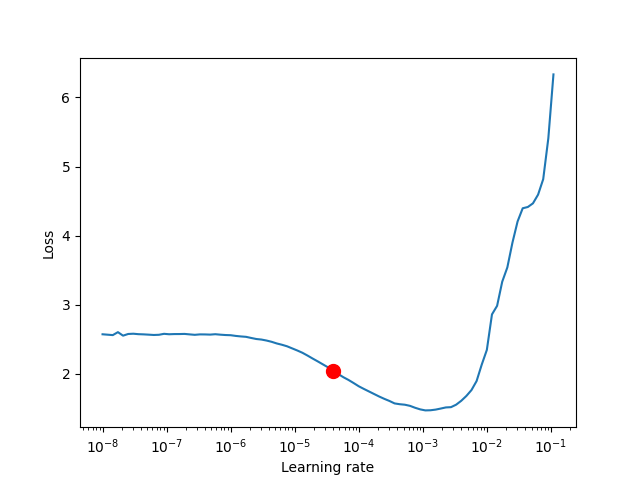}
    \caption{Learning rate finder:
        The dot represents the optimal learning rate for the neural network. 
        The optimal learning is not the learning rate that corresponds to the smallest loss, but the learning rate that corresponds to the greatest change in the loss, that is, the optimal learning is where the negative of the derivative is greatest. 
        }
\end{center}
\end{figure}

The authors did experiments with a learning rate finder. 
Overall, using a learning rate finder was worse for operator learning. 
The found learning rate was inconsistent and normally was not close to the optimal learning rate. 
Furthermore, experiments were done where the model was first warmed up before applying the learning rate finder, that is, a few epochs were done beforehand, and using a learning rate finder was still not effective.

%% file: conclusion.tex
For operator learning, it recommended to use $\mathrm{gelu}$ as the activation function. 
It is recommended to not use dropout. 
It is recommended to use stochastic weight averaging with a learning rate that is at most the original learning rate. 
And, it is recommended to not use a learning rate finder, but instead find it when hyperparameter tuning. 

%% file: code.tex
The code for the numerical experiments is publicly available at \url{https://gitlab.com/dustin_lee/neural-operators}.
All experiments were done in Python\index{Python}. 
Neural networks were implemented with PyTorch\index{PyTorch} \cite{paszke2017automatic, pointer2019programming, lippe2024uvadlc} and Lightning\index{Lightning} \cite{lightning}. 
Hyperparameter configuration was done using Hydra\index{Hydra} \cite{Yadan2019Hydra}.

%% file: competing-interest.tex
The authors declare that they have no known competing financial interests or personal relationships that could have appeared to influence the work reported in this paper.